\documentclass[11pt]{article}
\usepackage{booktabs}
\usepackage{CJKutf8}
\usepackage{amssymb}
\usepackage[russian,english]{babel}
\usepackage{tempora} 
\usepackage{eamt24}
\usepackage{times}
\usepackage{url}
\usepackage{latexsym}
\usepackage{arydshln}
\usepackage[small,bf]{caption} 
\title{Promoting Target Data in Context-aware Neural Machine Translation}

\author{
\\
\And
\\
\And
Harritxu Gete$^{1,2}$\\
  \And
\\
$^{1}$Vicomtech Foundation, Basque Research and Technology Alliance (BRTA)\\
$^{2}$University of the Basque Country UPV/EHU\\
{\tt \{hgete,tetchegoyhen\}@vicomtech.org}
 \And
Thierry Etchegoyhen$^{1}$\\
\And \\
\And \\
}

\date{}

\begin{document}
\maketitle

\begin{abstract}
 Standard context-aware neural machine translation (NMT) typically relies on parallel document-level data, exploiting both source and target contexts. Concatenation-based approaches in particular, still a strong baseline for document-level NMT, prepend source and/or target context sentences to the sentences to be translated, with model variants that exploit equal amounts of source and target data on each side achieving state-of-the-art results. In this work, we investigate whether target data should be further promoted within standard concatenation-based approaches, as most document-level phenomena rely on information that is present on the target language side. We evaluate novel concatenation-based variants where the target context is prepended to the source language, either in isolation or in combination with the source context. Experimental results in English-Russian and Basque-Spanish 
 show that including target context in the source leads to large improvements on target language phenomena. On source-dependent phenomena, using only target language context in the source achieves parity with state-of-the-art concatenation approaches, or slightly underperforms, whereas combining source and target context on the source side leads to significant gains across the board.
\end{abstract}

\section{Introduction}

Significant progress has been achieved in Machine Translation within the Neural Machine Translation (NMT) paradigm \cite{sutskever2014sequence,BahdanauCB14,VaswaniSPUJGKP17}. For the most part though, most NMT models translate sentences in isolation, preventing the adequate translation on document-level phenomena such as cohesion, discourse coherence or intersentential anaphora resolution \cite{bawden-etal-2018-evaluating,laubli-etal-2018-machine,voita-etal-2019-good,lopes-etal-2020-document,post2023escaping}.
Among the various approaches to context-aware NMT, simple concatenation of context sentences, as initially proposed by Tiedemann and Scherrer~\shortcite{tiedemann-scherrer-2017-neural}, remains a solid baseline typically used in practice with varying amounts of source-target context pairs \cite{Agrawal2018ContextualHI,junczys-dowmunt-2019-microsoft,majumder2022baseline,sun-etal-2022-rethinking,post2023escaping}.  

Context-aware models typically rely on parallel document-level data, a scarce resource overall despite recent efforts to provide this type of resource \cite{barrault-etal-2019-findings,voita-etal-2019-good,gete-etal-2022-tando}. To the exception of approaches such as the monolingual repair framework of Voita et al.~\shortcite{voita-etal-2019-context}, context data in the source language is generally used as the core information to model context-awareness. However, most discourse-level phenomena feature information that is either present mainly in the target language (e.g., lexical cohesion, deixis) or in both the source and target languages (e.g., gender selection, ellipsis). Considering this, in this work we aim to explore the impact of promoting target language data in standard context-aware NMT.

Along these lines, we explore a simple concatenation-based approach which consists in simply prepending context sentences from the target language to the source sentence to be translated, in isolation or in combination with source context. The underlying intuition is that contextual phenomena would be mainly modelled at the decoder level via target-side context information, whereas, on the encoder side, context data will be either ignored and copied, as foreign data, or also associated with source information to further model context. Using target language context data on the source side also enables the use of a standard NMT architecture and concatenation-based approach to context-aware NMT. 

We show that replacing source context sentences with the target context already leads to significant gains for discourse-level phenomena that depend on target-language information, while achieving either parity or moderate degradation in contrastive accuracy on other phenomena. Combining both source and target context sentences on the source side leads to consistent significant improvements across the board. We establish our results on two language pairs, English-Russian and Basque-Spanish, for which contrastive test sets are publicly available on a range of phenomena that depend on the source and/or target language context. 

In addition to accuracy results on specific phenomena, we compare the overall translation quality on parallel test sets as well. We also measure the impact of using either reference or machine-translated output as context at inference time, with only minor loss observed with the latter in our experiments. Finally, we evaluate the use of back-translated data, with similar comparative gains as those obtained using parallel document-level data.  Overall, our experimental results indicate that promoting target context data within a standard NMT architecture can be a promising alternative for context-aware machine translation.

\section{Related Work}
\label{sec:related-work}


One of the first methods proposed for document-level NMT is the concatenation of context sentences to the sentence to be translated, in either the source language only, or in both source and target languages \cite{tiedemann-scherrer-2017-neural,Agrawal2018ContextualHI}. This method does not require any architectural change and uses a fixed contextual window of sentences. It provides a robust baseline that often achieves performances comparable to that of more sophisticated methods, in particular in high-resource scenarios \cite{lopes-etal-2020-document,sun-etal-2022-rethinking,post2023escaping}. Variants of this approach include discounting the loss generated by the context \cite{lupo-etal-2022-focused}, extending model capacity \cite{majumder2022baseline,post2023escaping} or encoding the specific position of the context sentences \cite{lupo-etal-2023-encoding,gete-etchegoyhen-2023-evaluation}.

Alternative approaches include refining context-agnostic translations \cite{voita-etal-2019-context,mansimov-etal-2021-capturing} and modelling context information with specific NMT architectures \cite{jean-etal-2017-neural,li-etal-2020-multi-encoder,bao-etal-2021-g}. More recently, the use of pretrained language models has been explored for the task, using them to encode the context \cite{DBLP:journals/ml/WuXZWXQ22}, to initialize NMT models \cite{huang2023does} or fusing the language model with a sentence-level translation model \cite{petrick-etal-2023-document}. Directly using pretrained language models to perform translation can achieve competitive results, although these models might still produce critical errors and sometimes perform worse than conventional NMT models \cite{wang2023documentlevel,karpinska2023large,hendy2023good}. 

\begin{table*}[th]
\centering
\begin{tabular}{l}
\toprule
\textit{(a)} Lexical cohesion: name translation \\
\midrule
EN: Not for Julia. Julia has a taste for taunting her victims. \\
RU: \begin{otherlanguage*}{russian}Не для \textbf{Джулии}[\textit{Julia}]. \textbf{Юлия*}[\textit{Julia}] умеет дразнить своих жертв.\end{otherlanguage*}\\
\midrule
\textit{(b)} Deixis: register coherence \\
\midrule
EU: Ez dago martetarrik zuen artean. Guztiak ari zarete ereduak lotu eta...\\
ES: Ninguno de \textbf{ustedes}{[}form{]} es marciano. Todos \textbf{vosotros estáis*}{[}inf{]} siguiendo un modelo y...\\ 
\textit{(None of you are Martians. You are all following a model and...)}\\
\midrule
\textit{(c)} Gender selection \\ \midrule
EU: Hori nire \textbf{arreba} da. \textbf{Berak}{[}?{]} zaindu zituen nire argazkiak.   \\
\textit{(That’s my \textbf{sister}. \textbf{He/She} took care of my photos.)} \\
ES: Esa es mi \textbf{hermana}. \textbf{Él}* cuido mis fotos.  \\
\textit{(That’s my \textbf{sister}. \textbf{He*} took care of my photos.)}  \\\midrule
\textit{(d)} Verb phrase ellipsis\\ \midrule
EN: Veronica, thank you, but you \textbf{saw} what happened. We all \textbf{did}[?]. \\
RU: \begin{otherlanguage*}{russian}Вероника, спасибо, но ты \textbf{видела}, что произошло. Мы все \textbf{хотели}*.\end{otherlanguage*}\\
\textit{(Veronica, thank you, but you \textbf{saw} what happened. We all \textbf{wanted*} it.)}\\ 
\bottomrule
\end{tabular}
\caption{Examples of document-level inconsistencies extracted from \protect\cite{voita-etal-2019-good} and \protect\cite{gete-etal-2022-tando}.}
\label{tab:ex2}
\end{table*}

Concatenation-based approaches vary regarding their use of context, exploiting either the source context \cite{zhang-etal-2018-improving,voita-etal-2018-context}, the target context \cite{voita-etal-2019-context} or both \cite{bawden-etal-2018-evaluating,Agrawal2018ContextualHI,XuLW0C21,majumder2022baseline}. The benefits of using context sentences in both the source and the target languages are also discussed in Müller et al.~\shortcite{muller-etal-2018-large}, for a multi-encoder approach.
Fernandes et al. \shortcite{fernandes-etal-2021-measuring} conclude that concatenation-based models make more use of the target context than the source context, but Jin et al. \shortcite{jin-etal-2023-challenges} show that the effectiveness of the target context versus the source context is highly dependent on the language pair involved.
Close to the target-based approach we explore in this work, Scherrer et al.~\shortcite{scherrer-etal-2019-analysing} and Gete et al.~\shortcite{gete-etal-2023-works} include variants where target data is concatenated to the source sentence, notably showing that the target context is equally as important than source context, and particularly beneficial to address target-level phenomena. However, their experiments were limited to one target sentence, i.e. without prepending context on the target side. We show in this work that including the target context in both source and target languages is critical to achieve significant improvements overall.

Since standard NMT evaluation metrics such as BLEU \cite{papineni-etal-2002-bleu} are not well equipped to assess accuracy on discourse phenomena, several challenge test sets have been developed specifically to measure translations in context, via contrastive evaluations \cite{bawden-etal-2018-evaluating,muller-etal-2018-large,voita-etal-2019-good,lopes-etal-2020-document,nagata-morishita-2020-test,gete-etal-2022-tando,currey-etal-2022-mt}. We include contrastive test sets that cover target-language phenomena such as deixis or lexical cohesion, as well as phenomena where the relevant context information is available in both source and target languages.

\section{Exploiting Target Language Context}
\label{expl-tgt-data}

\begin{table*}[th]
\centering
\begin{tabular}{ll}
\toprule
\textit{(a)} & ES: Hablé con mi \textbf{amiga}{[}fem{]}. Dijo que sí.\\
& EN: I talked to my \textbf{friend}{[}?{]}. \textbf{She/He*} said yes.\\ \midrule
\textit{(b)} & EN: \textbf{You} can’t leave me! Don’t go away!\\
& ES: ¡No \textbf{puede} dejarme! ¡No \textbf{se vaya/te vayas*}!\\
\bottomrule
\end{tabular}
\caption{Example of ambiguity where source context is necessary for disambiguation, in isolation (a) or in combination with the target context (b).}
\label{tab:ex_source}
\end{table*}

The main incentive for the promotion of target context data is the nature of the contextual phenomena of interest for machine translation, as these can be grouped into four broad categories depending on the location of the relevant contextual information.

In a first category would be discourse-level phenomena that require context information on the target language side, typically related to discursive cohesion in a broad sense (see examples \textit{a} and \textit{b} in Table~\ref{tab:ex2}). For instance, to maintain lexical cohesion beyond the sentence level, a quality translation should feature lexical repetition when necessary, as it can mark emphasis or support question clarification. Another case is that of names with several possible translations, where translations must remain consistent throughout.  Degrees of politeness and linguistic register in general also involve translation alternatives that are equally correct in isolation, but require consistency at the document level. In the case of pronouns, when the source antecedent has translation options in different grammatical genders, translation choices should be coherent throughout in the target language. In all of these cases, the relevant information involves previous translations in the target language. 

In a second major category are phenomena for which the relevant context information is in both the source and the target context (examples \textit{c} and \textit{d} in Table~\ref{tab:ex2}). This includes word sense disambiguation scenarios, where different types of source or target elements may be relevant to perform disambiguation. Gender selection would also fall into this category, in those cases where translation options for the relevant contextual antecedent are unique or share the same gender. The resolution of elliptical constructions in the source language, with no equivalent in the target language, may also require context information from the source or the target language. Another instance for this type of phenomena would be the translation of Japanese zero pronouns into English \cite{nagata-morishita-2020-test}, where information on both sides can be relevant to determine the grammatical features of the target pronoun. Note that, even when contextual information is present in both the source and target languages, using  source information for disambiguation can result in a lack of consistency in the target language, whenever incorrect translations are involved.\footnote{Bawden et al.~\shortcite{bawden-etal-2018-evaluating} provide a contrastive test for these cases, where part of the source has been translated incorrectly but the translation is still required to be consistent overall.}

A third class of context-dependent phenomena exists, where source data are the only source of disambiguating information. This involves cases where the context includes the translation of a word marked for a specific category (e.g., gender) into a unmarked one, while the source sentence to be translated involves insufficient source information (e.g., a dropped pronoun) that needs to be translated into a marked element (e.g., a pronoun marked for gender). A typical example is provided in Table~\ref{tab:ex_source} \textit{a}. 
In such a case, there would be insufficient information in the target language, as the proper translation of the dropped subject pronoun into \textit{she} could only be determined from the gender of the source context antecedent \textit{amiga} (\textit{friend)}.

Finally, a fourth broad category contains constructions where the source and target context need to be processed in combination for a correct translation. In the example \textit{b} in Table~\ref{tab:ex_source}, the source context subject \textit{you} does not provide information about register, and neither does the target context in Spanish, since the verb \textit{puede} can indicate either third person in informal register or second person in polite register. However, the source context indicates second person. Therefore combining both sources of context information, it can be derived that the translation should be second person in polite form.


Any target-only approach, such as monolingual repair \cite{voita-etal-2019-context} or the target-only variant  we also explore in this work, would only generate the correct translation in the latter two classes of cases by either chance or training bias. Although these cases exist, it is unclear how widespread they actually are, compared to the other two main classes of contextual phenomena described above. In what follows, we set to compare the relative importance of source and target data across the main phenomena as represented in the selected document-level test suites. 

\section{Promoting Target Language Data}

To explore the promotion of target language data, we simply prepend the target context sentences to the source sentence to be translated, either discarding or maintaining the source context sentences. On the target side, we evaluate the use of empty context as well as maintaining the target context sentences. We add a special token to separate the concatenated context sentences in all cases. 

At inference time, in practice the previously translated sentences would be prepended as context. Since context translations can feature various degrees of correctness, we assess the approach under both ideal and average conditions. On parallel test sets, we measure the use of both correct reference context sentences (Section~\ref{sec:parallel-results-ref}) and machine-translated ones (Section~\ref{sec:mt-ref}). On the contrastive test sets, only reference translations are used, as is standard practice, since target context coherence requirements prevent the use of non-reference context translations for fair evaluations (see the discussion in Section~\ref{sec:mt-ref}). 

The prepended target-language data will need to be processed by the source language encoder under this approach, which might generate unwarranted noise. We hypothesise however that the encoder will essentially treat foreign language subwords as tokens to be copied directly into the target language, a typically simple operation for standard NMT models. We use  BPE models jointly learned on merged source and target language data to facilitate this part of the process. Overall, the proposed approach provides the means to exploit target language data on the decoder side, without any change to model architecture, while introducing data that might be easily processed via copying on the source side.  

\section{Experimental Setup}

\subsection{Data}

We describe in turn below the datasets used to train and test our models. All selected datasets were normalised, tokenised and truecased using Moses \cite{koehn2007moses} and segmented with BPE \cite{sennrich2016bpe}, training a joint model over 32,000 operations. Tables~\ref{corp:parallel} and~\ref{corp:contrastive} show corpora statistics for parallel and contrastive datasets respectively.

\begin{table}[th]
\centering
\begin{tabular}{lrr}
\toprule
 & \multicolumn{1}{c}{EU-ES} & \multicolumn{1}{c}{EN-RU} \\ \midrule
\textsc{Train}& 1,753,726  & 6,000,000 \\
\textsc{Dev}   & 3,051 & 10,000 \\
\textsc{Test}  & 6,078 & 10,000 \\ \bottomrule
\end{tabular}
\caption{Parallel corpora statistics (number of sentences)}
\label{corp:parallel}
\end{table}

\begin{table}
\centering
\begin{tabular}{lrcccc}
\toprule

EU-ES & \multicolumn{1}{c}{Size} & src & tgt & Dist. \\
                 
\midrule
GDR-SRC+TGT    & 300                                    & \checkmark   & \checkmark & $\leq5$  \\
COH-TGT          & 300                                    &     & \checkmark   & $\leq5$\\

\midrule
EN-RU & \multicolumn{1}{l}{Size} & src & tgt & Dist. \\

\midrule
Ellipsis (infl.) & 500                                    & \checkmark   & \checkmark   & $\leq3$\\
Ellipsis (VP)    & 500                                    & \checkmark   & \checkmark   & $\leq3$\\
Deixis           & 2,500                                   &     & \checkmark  & $\leq3$ \\
Lex. cohesion    & 1,500                                   &     & \checkmark & $\leq3$ \\
\bottomrule
\end{tabular}
\caption{Contrastive test sets: size (number of instances), required context information and distance to the disambiguating information (number of sentences)}
\label{corp:contrastive}
\end{table}

For Basque--Spanish, we selected the TANDO corpus \cite{gete-etal-2022-tando}, which contains parallel data from subtitles, news and literary documents. It includes two contrastive datasets for Basque to Spanish translation. The first one, GDR-SRC+TGT, centres on gender selection, with the disambiguating information present in both the source and target languages. The second one, COH-TGT, is meant to evaluate cases where, despite the absence in the source language of the necessary information to make a correct selection of gender or register, the translation must be contextually coherent using target-side information.

For English--Russian, we used the dataset described in Voita et al.~\shortcite{voita-etal-2019-good}, based on Open Subtitles excerpts \cite{lison-etal-2018-opensubtitles2018}. It includes 4 large-scale contrastive test sets for English to Russian translation.
Two of these tests are related to ellipsis and contain the disambiguating information in both the source and target-side context: Ellipsis infl. assesses the selection of correct morphological noun phrase forms in cases where the source verb is elided, whereas Ellipsis VP evaluates the ability to predict the verb in Russian from an English sentence in which the verb phrase is elided. In the other two tests, the disambiguating information is only present in the target-side context: Deixis addresses politeness consistency in the target language, without nominal markers, whereas Lexical Cohesion focuses on the consistent translation of named entities in Russian.  

\subsection{Models}


All models in our experiments are trained with Marian \cite{junczys-dowmunt-etal-2018-marian} and rely on the Transformer-base architecture with the parameters described in Vaswani et al.~\shortcite{VaswaniSPUJGKP17}. 

As a general baseline, we trained a sentence-level model using all source-target sentence pairs in the selected training datasets for each language pair. We then trained different variants of concatenation-based context-aware models, varying the type of context sentences prepended to the source and/or the target sentence, and adding a special token to separate the context.

We use the following convention to denote the models: $n$to$n$ uses the same amount of source and target data on each side, and represents the state-of-the-art baseline; tgt-$n$to$n$ uses target language data on both sides, discarding source context altogether; $n$to$1$ and tgt-$n$to$1$ are variants of the previous models that use no context sentences in the target language; finally, src+tgt-$n$to$n$ and tgt+src-$n$to$n$ are variants where target context sentences are combined with source context sentences, by prepending them after or before the latter, respectively. For convenience, we will refer to the tgt-$n$to$n$, src+tgt-$n$to$n$ and tgt+src-$n$to$n$ variants as X-tgt-$n$to$n$, as they share the use of target context on both sides. 

Given the size of the context for each dataset, we have $n$=6 for Basque--Spanish models and $n$=4 for English--Russian models. All context-aware models were initialised with the weights of the sentence-level baseline.

Note that we discarded $1$to$n$ models, as they present two main challenges. Within a standard concatenation approach, we would be tasking the model to learn a transformation from a single source sentence to both the context and the target sentence, although the target context cannot be derived from the source sentence, obviously. Alternatively, a $1$to$n$ model could be designed via changes in the NMT architecture, with forced decoding over the specified target context at both training and inference time. The required architectural changes were beyond the scope of this work, although this type of model might be worth exploring in more details. 


        

\section{Results}
\label{sec:results}

\subsection{Parallel Tests}
\label{sec:parallel-results-ref}

We first compared models in terms of BLEU on the parallel test sets, using SacreBLEU \cite{post-2018-call}\footnote{nrefs:1|case:mixed|eff:no|tok:13a|smooth:exp|version:2.3.1}. Statistical significance was computed via paired bootstrap resampling \cite{koehn-2004-statistical}, for $p<0.05$.\footnote{In all tables, best scores given the statistical test at hand are shown in bold; significantly better results between $n$to$n$ and tgt-$n$to$n$ results are indicated with \dag.} The results are shown in Table~\ref{results:bleu}. 

In Basque--Spanish, the $n$to$n$, tgt-$n$to$n$, and src+tgt-$n$to$n$ models performed better than the alternatives, with no statistically significant differences between the three, with the tgt+src-$n$to$n$ achieving slightly lower results. All three were notably significantly better than the baseline and the models which used only a single reference in the target language. In English--Russian, all $X$-tgt-$n$to$n$ model variants, that included target context data on the source side, outperformed all other models, including the standard $n$to$n$ model.

\begin{table}[th]
\centering
\begin{tabular}{lcc}
\toprule
 & \multicolumn{1}{c}{EU-ES} & \multicolumn{1}{c}{EN-RU}  \\
 \midrule
Sentence-level  & 31.20 & 31.09 \\
$n$to$1$ & 29.91 &  31.48  \\
tgt-$n$to1        & 29.43 & 31.03 \\
$n$to$n$            & $\textbf{31.96}$ &  31.20 \\
tgt-$n$to$n$& $\textbf{31.82}$ &  $\textbf{32.29}$ \\            
src+tgt-$n$to$n$ & $\textbf{31.94}$ &  $\textbf{32.32}$  \\
tgt+src-$n$to$n$ & 31.56 & $\textbf{32.49}$   \\ 
\bottomrule
\end{tabular}
\caption{BLEU results on the parallel test sets.}
\label{results:bleu}
\end{table}

Sentence-level metrics are typically insufficient to assess translation quality at the document level \cite{wong-kit-2012-extending}, and conclusions should not be drawn from the above results regarding context-aware ability of the different models. They do however indicate several tendencies at the sentence level. First, the proposed use of target context data on both sides was not detrimental in terms of translation quality, as the X-tgt-$n$to$n$ models performed on a par with, or better than, the other variants. Secondly, the lower results obtained by the $n$to1 and tgt-$n$to1 models seem to indicate that (i) removing target context data on the decoder side can be detrimental, as in EU-ES, and (ii) using source or target language data on the encoder side can lead to similar BLEU results, as was the case in both language pairs. 

Note that the results above were obtained with reference translations, in an ideal scenario where the context is correctly translated. In Section~\ref{sec:mt-ref}, we present additional results using machine-translated context, to measure the impact of eventual errors in target context translation. 

\subsection{Challenge Tests}
\label{sec:challenge-tests}

We evaluated the different models on the challenge test sets both in terms of BLEU and in terms of accuracy of the contrastive evaluation. Statistical significance of accuracy results was computed using McNemar’s test \cite{Mcnemar1947NoteOT}, for $p<0.05$. 
The results are shown in Tables~\ref{results:acc_eseu} and~\ref{results:acc_enru}. 


\begin{table*}[]
\centering
\begin{tabular}{lccccc}
\toprule
               & \multicolumn{2}{c}{GDR-SRC+TGT}                     && \multicolumn{2}{c}{COH-TGT}     \\ \cmidrule{2-3} \cmidrule{5-6}
               & \multicolumn{1}{l}{BLEU} & \multicolumn{1}{l}{ACC.} && \multicolumn{1}{l}{BLEU} & ACC. \\
               \midrule
Sentence-level & 36.28 & 53.67 && 35.04 & 54.00 \\
$n$to$1$           & 36.82 & 66.33 && 33.23& 53.00 \\
tgt-$n$to$1$       & 36.79 & 66.33 && 37.31 & 74.00 \\
$n$to$n$           & 40.45 & $\textbf{77.67}$ && 35.89 & 65.33 \\
tgt-$n$to$n$       & $39.05$ & 72.67 && $\textbf{39.61}$ & $\textbf{81.67}$ \\
src+tgt-$n$to$n$    & $\textbf{41.29}$ & $\textbf{78.67}$ && $\textbf{40.23}$ & $\textbf{84.67}$  \\
tgt+src-$n$to$n$    & $\textbf{42.35}$ & $\textbf{78.67}$ && $\textbf{39.86}$ & $\textbf{82.67}$  \\
\bottomrule
\end{tabular}
\caption{BLEU and accuracy results on the Basque--Spanish challenge tests.}
\label{results:acc_eseu}
\end{table*}

\begin{table*}[]
\centering
\begin{tabular}{lccccccccccc}
\toprule
               & \multicolumn{2}{c}{Ellipsis infl.} && \multicolumn{2}{c}{Ellipsis VP} && \multicolumn{2}{c}{Deixis} && \multicolumn{2}{c}{Lex. Cohesion} \\ \cmidrule{2-3} \cmidrule{5-6} \cmidrule{8-9} \cmidrule{11-12} 
               & \multicolumn{1}{l}{BLEU} & \multicolumn{1}{l}{ACC.} && \multicolumn{1}{l}{BLEU} & ACC. && \multicolumn{1}{l}{BLEU} & \multicolumn{1}{l}{ACC.} && \multicolumn{1}{l}{BLEU} & ACC.\\
               \midrule
Sentence-level & 30.81 & 51.80 &  & 22.20 & 27.80 &  & 28.10 & 50.04 &  & \textbf{31.52} & 45.87 \\
$n$to$1$           & 32.69  & 54.60 &  & 30.24  & 65.40 &  & 28.20  & 50.04 &  & 29.47  & 45.87 \\
tgt-$n$to$1$       &  32.28 & 53.60 &  & 23.59 & 29.00 &  & 28.30 & 50.56 &  & 30.37 & 45.87 \\
$n$to$n$           & 36.97 & $\textbf{75.20}$ &  & 29.59 & 62.60 &  & 27.15 & 82.48 &  & 27.89 & 45.93 \\
tgt-$n$to$n$       & $\textbf{40.69}$ & 70.00   &  & 30.75 & $60.00$   &  & $\textbf{34.17}$ & $\textbf{87.48}$ &  & 30.98 & 49.47 \\
src+tgt-$n$to$n$ & $\textbf{40.98}$ & \textbf{77.20}   && \textbf{35.84}  & \textbf{77.60} && \textbf{34.38} & $\textbf{87.48}$ && $\textbf{31.75}$ & $\textbf{53.07}$  \\
tgt+src-$n$to$n$ & \textbf{42.02} & \textbf{75.60} && \textbf{34.46} & \textbf{74.88} && \textbf{34.07} & \textbf{88.28} && \textbf{31.33} & 51.00  \\
\bottomrule
\end{tabular}
\caption{BLEU and accuracy results in English--Russian challenge tests. }
\label{results:acc_enru}
\end{table*}

Considering both language pairs, the first notable results are the significant gains achieved by the src+tgt-$n$to$n$ and tgt+src-$n$to$n$ models, which outperformed all other variants overall, in terms of both BLEU scores and contrastive accuracy. The tgt-$n$to$n$ model, where source context was discarded altogether, also outperformed the baselines in terms of BLEU in all but one case, and either matched the other two target-based variants in half of the  scenarios, or  was outperformed by these variants in the other three cases. In terms of contrastive accuracy, it also outperformed the baselines by a wide margin on target-oriented phenomena while achieving parity or resulting in accuracy loss on other phenomena. Overall, the best performing and most consistent variant across datasets and metrics was the src+tgt-$n$to$n$ variant.

On all target-related phenomena, the X-tgt-$n$to$n$ models outperformed all alternatives, and in particular the standard $n$to$n$ variant by large margins. In terms of accuracy, in EU-ES on the COH-TGT test, the tgt-$n$to$n$ model already outperformed the baseline by 27.67 points and the $n$to$n$ model by 16.34 points, with even higher accuracy gains for the best-performing src+tgt-$n$to$n$ model (+19.34). In EN-RU, on Deixis gains of up to 38.24 and 5.8 points were achieved against the baseline and $n$to$n$ model, respectively; on the Lexical Cohesion test set, the gains reached 7.2 and and 7.14 points, respectively.
On these target-oriented test-sets, all X-tgt-$n$to$n$ model also achieved comparable gains in terms of BLEU scores, with a maximum against the $n$to$n$ model of +4.34 points in EU-ES, +7.23 in EN-RU on Deixis, and +3.86 in EN-RU on the Lexical cohesion test.

Turning now to the test sets where relevant context information is available in either both the source and target languages, or perhaps only in the source language in some cases, the results are more balanced between the $n$to$n$ baseline and the X-tgt-$n$to$n$ variants, although the src+tgt-$n$to$n$ achieved the best results overall in terms of both BLEU and accuracy. On Ellipsis VP, the latter notably achieved gains of 15 accuracy points, with the tgt+src-$n$to$n$ variant a close second at +12.28. On Ellipsis infl. and GDR-SRC-TGT, the gains were more limited, with a maximum of +1 and and +2 accuracy points for the src+tgt-$n$to$n$ model against the $n$to$n$ baseline, respectively, although significant BLEU gains of up to +3.3 and +5.05 were observed on these test sets, respectively.

Unsurprisingly, on these three datasets where source information is a relevant factor, in combination or in isolation, the tgt-$n$to$n$ model underperformed, though in accuracy only and to a limited extended on Ellipsis VP, for instance. This variant also significantly outperformed the $n$to$n$ baseline in terms of BLEU on Ellipsis infl., with a 3.60 points gain. To further determine the impact of source and target context and more precisely asses the limits of this type of model, more fine-grained challenge tests would be needed to distinguish between cases that can solely be resolved with source context information and those where either side of context provides sufficient information. 

Regarding the other two contextual variants, $n$to$1$ and tgt-$n$to$1$, which used no context information on the target side of the input, the results in accuracy were similar overall, performing on a par with the sentence-level baseline on Lexical Cohesion, Deixis and COH-TGT for $n$to$1$. This was expected for the $n$to$1$ models, as the relevant information is in the target language in these cases, which these models have no access to.


Overall, promoting target data in a concatenation-based approach achieved large improvements across the board over the sentence-level and $n$to$n$ baselines. Replacing source context data altogether with the target context already improved significantly on target-context phenomena, while achieving relatively close results in the other cases. Combining source and target context provided the best balance however, achieving the best results in all cases. In particular, the src+tgt-$n$to$n$ proved optimal and we discarded the slightly worse tgt+src-$n$to$n$  variant in the remainder of this work.

\section{Using Back-translated Data}
\label{sec:bt}

When document-level parallel data are lacking, monolingual data in the target language can be exploited within concatenation-based approaches via back-translation \cite{junczys-dowmunt-2019-microsoft,sugiyama-yoshinaga-2019-data,huo-etal-2020-diving}. Some level of degradation is expected, depending on the quality of the model used to back-translate the target data, and we also expect the models to be impacted differently: the target sentence and its back-translation would be identical for all models, as would be the original target context sentences, but the $n$to$n$ and the src+tgt-$n$to$n$ models also require back-translated context, unlike the tgt-$n$to$n$ model. 

For comparison purposes we back-translated the target side of the training data for both language pairs, using a sentence-level model trained on the parallel data, and trained the main model variants strictly on the back-translated data.\footnote{Note that we did not mix back-translated data with the original parallel data, to strictly contrast the approaches in their ability to exploit monolingual back-translated data.} The results are shown in Table~\ref{results:bleu_bt}, contrasting the use of parallel (PA) and back-translated (BT) data. 

The overall degradation using BT data was more salient in EU-ES than in EN-RU,  which is likely due to the differences in training data size and the resulting quality of the respective models. In both cases, the X-tgt-$n$to$n$ variants proved more robust than the $n$to$n$ model. This is also likely due to the latter having as context only the back-translation of the target context, while the former contain, alone or in combination with the back-translation, the original target context.


\begin{table}[th]
\centering
\begin{tabular}{lcc}
\toprule
  & \multicolumn{1}{l}{EU-ES} & \multicolumn{1}{l}{EN-RU}  \\
 \midrule
Sentence-level (PA)  & 31.20 & 31.09 \\ \midrule
$n$to$n$ (PA) & $\textbf{31.96}$ & 31.20 \\
tgt-$n$to$n$ (PA) & $\textbf{31.82}$ & \textbf{32.29} \\
src+tgt-$n$to$n$ (PA) & $\textbf{31.94}$ & \textbf{32.32}  \\ \midrule     
$n$to$n$ (BT) & 25.46& 29.21\\
tgt-$n$to$n$ (BT) & 27.33 & \textbf{30.10}\\   
src+tgt-$n$to$n$ (BT) & \textbf{31.27} & 29.39 \\
\bottomrule
\end{tabular}
\caption{BLEU results on the parallel test sets using  parallel (PA) and back-translated (BT) data.}
\label{results:bleu_bt}
\end{table}

\begin{table*}[]
\centering
\begin{tabular}{lcccccc}
\toprule
 & \multicolumn{2}{c}{GDR-SRC+TGT}                     && \multicolumn{2}{c}{COH-TGT}     \\ \cmidrule{2-3} \cmidrule{5-6}
               & \multicolumn{1}{l}{BLEU} & \multicolumn{1}{l}{ACC.} && \multicolumn{1}{l}{BLEU} & ACC. \\
               \midrule
Sentence-level          & 36.28 & 53.67 && 35.04 & 54.00 \\ \midrule
$n$to$n$ (PA)           & \textbf{40.45} & \textbf{77.67} && 35.89 & 65.33 \\
tgt-$n$to$n$ (PA)       & 39.05 & 72.67 && \textbf{39.61} & \textbf{81.67} \\
src+tgt-$n$to$n$ (PA)   & \textbf{41.25} & \textbf{78.67} && \textbf{40.23} & \textbf{84.67}  \\ \midrule
$n$to$n$ (BT)           & 41.58 & \textbf{76.00} && 31.02& 67.00\\
tgt-$n$to$n$ (BT)       & 40.22 & 74.00 && 34.62 & \textbf{81.33} \\
src+tgt-$n$to$n$ (BT)   & \textbf{45.67} & \textbf{77.33} && \textbf{42.67} & \textbf{84.67} \\
\bottomrule
\end{tabular}
\caption{Results on Basque--Spanish contrastive tests with parallel (PA) and back-translated (BT) data.}
\label{results:acc_eseu_bt}
\end{table*}

\begin{table*}[]
\centering
\begin{tabular}{lcccccccccccc}
\toprule
    & \multicolumn{2}{c}{Ellipsis infl.} && \multicolumn{2}{c}{Ellipsis VP} && \multicolumn{2}{c}{Deixis} && \multicolumn{2}{c}{Lex. cohesion} \\ \cmidrule{2-3} \cmidrule{5-6} \cmidrule{8-9} \cmidrule{11-12} 
               & \multicolumn{1}{l}{BLEU} & \multicolumn{1}{l}{ACC.} && \multicolumn{1}{l}{BLEU} & ACC. && \multicolumn{1}{l}{BLEU} & \multicolumn{1}{l}{ACC.} && \multicolumn{1}{l}{BLEU} & ACC.\\
               \midrule

Sentence-level          & 30.81 & 51.80 &  & 22.20 & 27.80 &  & 28.10 & 50.04 &  & 31.52 & 45.87 \\ \midrule
$n$to$n$ (PA)           & 36.97 & \textbf{75.20} & & 29.59 & 62.60 &  & 27.15 & 82.48 & & 27.89 & 45.93 \\
tgt-$n$to$n$ (PA)       & \textbf{40.69} & 70.00 & & 30.75 & 60.00 & & \textbf{34.17} & \textbf{87.48} & & 30.98 & 49.47 \\
src+tgt-$n$to$n$ (PA)   & \textbf{40.98} & \textbf{77.20} && \textbf{35.84} & \textbf{77.60} && \textbf{34.38} & \textbf{87.48} && \textbf{31.75} & \textbf{53.07} \\ \midrule
$n$to$n$ (BT)           & 35.63& \textbf{78.60} &  & 28.84& 69.40 & & 25.66& 83.92&  & 28.29& 46.20\\
tgt-$n$to$n$ (BT)       & \textbf{39.25} & 73.60 & & 31.86 & 57.60&  & \textbf{31.84} & \textbf{87.84} &  & 29.81 & 49.20 \\
src+tgt-$n$to$n$ (BT)   & \textbf{41.96} & \textbf{81.20} && \textbf{35.23} & \textbf{76.00} && \textbf{31.63} & \textbf{87.36} && \textbf{31.68} & \textbf{66.07}\\

\bottomrule
\end{tabular}
\caption{Results on English--Russian contrastive tests with  parallel (PA) and back-translated (BT) data.}
\label{results:acc_enru_bt}
\end{table*}



Overall, the tendencies observed using parallel data were replicated with back-translated data, with the src+tgt-$n$to$n$ model being the top-performing variant across the board, and the tgt-$n$to$n$ a close second on target-context phenomena but performing worse than the $n$to$n$ model in accuracy on the GDR-SRC+TGT and Ellipsis infl. with BT data. Perhaps more surprising are the results achieved by the src+tgt-$n$to$n$ model, trained on BT data, on the Lexical cohesion test set, where it outperformed the same variant trained on parallel data by 13 points. Additional datasets might be warranted to further assess the tendencies for these models, but the results on the available datasets in terms of accuracy seem to indicate that the use of BT data is viable, and particularly exploitable by the X-tgt-$n$to$n$ models overall. We conjecture that this is mainly due to the fact that these approaches promote target language data which are in essence correct, while discarding or reducing the role of source context data which are likely to feature back-translation errors.

\section{Machine-translated Target Context}
\label{sec:mt-ref}

Following standard practice, so far we have used the reference target context instead of the machine-translated output in our evaluations. This is meant to remove potential noise in terms of context translation errors and evaluate the approaches on their translation accuracy given a correct context.  Using reference translations also allows for an evaluation of phenomena where more than one context translation would be correct -- e.g. \textit{box} translated as \textit{boîte} (fem.) instead of \textit{carton} (masc.) in French -- but the contrastive evaluation relies on one of these translations being selected as the correct one and further phenomena, such as coherence, are measured accordingly. A correct but different context translation would unfairly affect the results in these cases.

\begin{table}[th]
\centering
\begin{tabular}{lccc}
\toprule
& \multicolumn{1}{l}{EU-ES} & \multicolumn{1}{l}{EN-RU}  \\
 \midrule
Sentence-level  & 31.20 & 31.09 \\
$n$to$n$     & \textbf{31.96} & 31.20 \\
tgt-$n$to$n$ (RF) & \textbf{31.82} & \textbf{32.29} \\
tgt-$n$to$n$ (MT) & 31.08 & 31.52\\ 
src+tgt-$n$to$n$ (RF) & \textbf{31.94} & \textbf{32.32} \\
src+tgt-$n$to$n$ (MT) & 30.93 & 31.31 \\ 
\bottomrule
\end{tabular}
\caption{BLEU results on the parallel test sets using  reference (RF) and machine-translated (MT) context.}
\label{results:bleu_mt}
\end{table}

\begin{table*}[]
\centering
\begin{tabular}{llrccclrccc}
\hline
\textbf{}     &&& \multicolumn{3}{c}{GDR-SRC+TGT}                                                                    & \multicolumn{1}{c}{\textbf{}} && \multicolumn{3}{c}{COH-TGT}                                                                        \\ \cline{4-6} \cline{9-11} 
\textit{Dist} &  & \% cases & \multicolumn{1}{c}{$n$to$n$} & \multicolumn{1}{c}{tgt-$n$to$n$} & \multicolumn{1}{c}{src+tgt-$n$to$n$} & \multicolumn{1}{c}{} & \% cases & \multicolumn{1}{c}{$n$to$n$} & \multicolumn{1}{c}{tgt-$n$to$n$} & \multicolumn{1}{c}{src+tgt-$n$to$n$} \\ \hline
1 && 64.67\% & 77.32 & 70.10 & 76.80 && 62.34\% & 69.52 & 85.03 & 86.10 \\
2 && 20.67\% & 91.23 & 85.48 & 85.48 && 20.67\% & 66.13 & 90.32 & 85.48 \\
3 && 9.33\% & 72.41 & 71.43 & 71.43 && 9.67\% & 51.72 & 72.41 & 75.86 \\
4 && 2.00\% & 57.14 & 57.14 & 85.71 && 6.00\% & 50.00 & 83.33 & 83.33 \\
5 && 3.33\% & 66.67 & 55.56 & 88.89 && 1.33\% & 25.00 & 50.00 & 75.00 \\ \hline
\end{tabular}
\caption{Accuracy results in Basque--Spanish according to relevant context distance.}
\label{results:acc_eues_dist}
\end{table*}

\begin{table*}[]
\centering
\begin{tabular}{llrccclrccc}
\hline
\textbf{}     &&  & \multicolumn{3}{c}{Deixis}                                                                     & \multicolumn{1}{c}{\textbf{}} && \multicolumn{3}{c}{Lex. Cohesion}                                                                  \\ \cline{4-6} \cline{9-11} 
\textit{Dist} &  & \% cases & \multicolumn{1}{c}{$n$to$n$} & \multicolumn{1}{c}{tgt-$n$to$n$} & \multicolumn{1}{c}{src+tgt-$n$to$n$} && \multicolumn{1}{c}{\% cases}          & \multicolumn{1}{c}{$n$to$n$} & \multicolumn{1}{c}{tgt-$n$to$n$} & \multicolumn{1}{c}{src+tgt-$n$to$n$} \\ \hline
1 & & 33.33\% & 88.66 & 90.49 & 89.63 && 42.75\% & 46.27 & 51.45 & 57.53 \\
2 & & 33.33\%  & 85.82 & 90.07 & 91.02 && 31.50\% & 45.87 & 47.39 & 50.00 \\
3 & & 33.33\%  & 73.02 & 81.89 & 81.77 && 25.75\% & 45.43 & 48.56 & 49.09 \\ \hline
\end{tabular}
\caption{Accuracy results in English--Russian according to relevant context distance. }
\label{results:acc_enru_dist}
\end{table*}

Still, in practice, at inference time there are no reference translations, of course. Whereas the $n$to$n$ model should not be impacted at all, 
the X-tgt-$n$to$n$ models are susceptible to suffer from errors in the translation of the context. To measure this aspect, we computed BLEU scores using machine-translated target sentences for X-tgt-$n$to$n$ models. The results are shown in Table~\ref{results:bleu_mt}. 

From these results, using MT output did not seem to markedly impact translation quality, at least in terms of BLEU scores. As previously noted, measuring its impact on contrastive accuracy would not lead to a fair evaluation, as these challenge sets rely on specific choices among the set of valid translations of the context.
Additionally, a proper assessment of the impact of machine-translated context on the X-tgt-$n$to$n$ models would need to take into account the quality of the translation model itself, with larger models expected to minimise context translation errors in this type of approach.

\section{Accuracy At Distance}



The results so far were measured considering context as a whole. To achieve a more fine-grained view of the differences between approaches, 
we computed their accuracy in terms of the distance between the current sentence and the disambiguating context information, expressed in number of sentences. The results are shown in Tables~\ref{results:acc_eues_dist} and \ref{results:acc_enru_dist}, indicating the distance and the percentages of cases in the corresponding dataset.

The main observable tendency is that of the decreasing accuracy over distance for the $n$to$n$ model, in all cases but GDR-SRC+TGT at distance 2 (where all models perform better), in contrasted with the significantly more robust accuracy of the src+tgt-$n$to$n$ model at larger distances, for Basque-Spanish in particular. The tgt-$n$to$n$ model exhibits mixed tendencies, improving or maintaining accuracy over distance 1 in some cases, but also degrading at larger distances (GDR-SRC+TGT or COH-TGT, at dist=5). Note though that larger distances are under-represented in the Basque-Spanish test sets, and may thus not be as representative.

\section{Conclusions}

In this work, we investigated the impact of promoting target context data within a standard concatenation-based approach to context-aware neural machine translation. The main incentive for this exploration revolves around the fact that, for most contextual phenomena of interest for document-level machine translation, the relevant information is either in the target language or distributed on the source and target sides. Although we also described a class of phenomena where only the source context can lead to an accurate translation, it is unclear how widespread these cases actually are overall.

To assess target data promotion, we proposed simple variants where target context sentences are simply concatenated to the source sentence, either in isolation or in combination with the source context. Our results in Basque-Spanish and English-Russian, over five datasets showcasing different types of contextual phenomena, show that large improvements can be obtained by promoting target context, in terms of contrastive accuracy and BLEU scores, notably on target-language phenomena. Models where the source context was discarded altogether achieved parity or slightly underperformed on context phenomena involving both source and target contexts. The variants based on augmenting the source context with target data achieved the best results across the board and were also shown to be more accurate in handling context at larger distances.

We further evaluated the use of back-translated data, showing that the tendencies observed on parallel data were maintained, with models trained on promoting target context in combination with source context matching or even outperforming models trained on parallel data.

We also measured the impact of using machine-translated output instead of reference translations, which could have impacted the proposed approach but were shown to have only a marginal effect, on the parallel test sets at least. The use of more robust baseline models, trained on larger volumes of data, should further mitigate these effects. New evaluation protocols could support a more precise evaluation of these aspects, addressing the current dependency of challenge datasets on arbitrary context translation decisions for some phenomena. 

The proposed approach promoting target data requires no changes to the standard NMT architecture and provides significant gains over strong baselines. Although it also implies larger contexts when using source and target context in combination on the source language side, it might be worth further exploring this type of approach and the respective roles of source and target context data  in neural machine translation.


\bibliography{eamt24,anthology}

\begin{thebibliography}{}

\bibitem[\protect\citename{Agrawal \bgroup et al.\egroup }2018]{Agrawal2018ContextualHI}
Agrawal, Ruchit, Marco Turchi, and Matteo Negri.
\newblock 2018.
\newblock Contextual handling in neural machine translation: Look behind, ahead and on both sides.
\newblock In {\em Proceedings of the 21st Annual Conference of the European Association for Machine Translation (EAMT)}.

\bibitem[\protect\citename{Bahdanau \bgroup et al.\egroup }2015]{BahdanauCB14}
Bahdanau, Dzmitry, Kyunghyun Cho, and Yoshua Bengio.
\newblock 2015.
\newblock Neural machine translation by jointly learning to align and translate.
\newblock In {\em 3rd International Conference on Learning Representations, {ICLR} 2015}.

\bibitem[\protect\citename{Bao \bgroup et al.\egroup }2021]{bao-etal-2021-g}
Bao, Guangsheng, Yue Zhang, Zhiyang Teng, Boxing Chen, and Weihua Luo.
\newblock 2021.
\newblock {G}-transformer for document-level machine translation.
\newblock In {\em Proceedings of the 59th Annual Meeting of the Association for Computational Linguistics and the 11th International Joint Conference on Natural Language Processing (Volume 1: Long Papers)}, pages 3442--3455, Online, August. Association for Computational Linguistics.

\bibitem[\protect\citename{Barrault \bgroup et al.\egroup }2019]{barrault-etal-2019-findings}
Barrault, Lo{\"\i}c, Ond{\v{r}}ej Bojar, Marta~R. Costa-juss{\`a}, Christian Federmann, Mark Fishel, Yvette Graham, Barry Haddow, Matthias Huck, Philipp Koehn, Shervin Malmasi, Christof Monz, Mathias M{\"u}ller, Santanu Pal, Matt Post, and Marcos Zampieri.
\newblock 2019.
\newblock Findings of the 2019 conference on machine translation ({WMT}19).
\newblock In {\em Proceedings of the Fourth Conference on Machine Translation (Volume 2: Shared Task Papers, Day 1)}, pages 1--61, Florence, Italy, August. Association for Computational Linguistics.

\bibitem[\protect\citename{Bawden \bgroup et al.\egroup }2018]{bawden-etal-2018-evaluating}
Bawden, Rachel, Rico Sennrich, Alexandra Birch, and Barry Haddow.
\newblock 2018.
\newblock Evaluating discourse phenomena in neural machine translation.
\newblock In {\em Proceedings of the 2018 Conference of the North {A}merican Chapter of the Association for Computational Linguistics: Human Language Technologies, Volume 1 (Long Papers)}, pages 1304--1313, New Orleans, Louisiana, June. Association for Computational Linguistics.

\bibitem[\protect\citename{Currey \bgroup et al.\egroup }2022]{currey-etal-2022-mt}
Currey, Anna, Maria Nadejde, Raghavendra~Reddy Pappagari, Mia Mayer, Stanislas Lauly, Xing Niu, Benjamin Hsu, and Georgiana Dinu.
\newblock 2022.
\newblock {MT}-{G}en{E}val: A counterfactual and contextual dataset for evaluating gender accuracy in machine translation.
\newblock In Goldberg, Yoav, Zornitsa Kozareva, and Yue Zhang, editors, {\em Proceedings of the 2022 Conference on Empirical Methods in Natural Language Processing}, pages 4287--4299, Abu Dhabi, United Arab Emirates, December. Association for Computational Linguistics.

\bibitem[\protect\citename{Fernandes \bgroup et al.\egroup }2021]{fernandes-etal-2021-measuring}
Fernandes, Patrick, Kayo Yin, Graham Neubig, and Andr{\'e} F.~T. Martins.
\newblock 2021.
\newblock Measuring and increasing context usage in context-aware machine translation.
\newblock In {\em Proceedings of the 59th Annual Meeting of the Association for Computational Linguistics and the 11th International Joint Conference on Natural Language Processing (Volume 1: Long Papers)}, pages 6467--6478, Online, August. Association for Computational Linguistics.

\bibitem[\protect\citename{Gete and Etchegoyhen}2023]{gete-etchegoyhen-2023-evaluation}
Gete, Harritxu and Thierry Etchegoyhen.
\newblock 2023.
\newblock An evaluation of source factors in concatenation-based context-aware neural machine translation.
\newblock In Mitkov, Ruslan and Galia Angelova, editors, {\em Proceedings of the 14th International Conference on Recent Advances in Natural Language Processing}, pages 399--407, Varna, Bulgaria, September. INCOMA Ltd., Shoumen, Bulgaria.

\bibitem[\protect\citename{Gete \bgroup et al.\egroup }2022]{gete-etal-2022-tando}
Gete, Harritxu, Thierry Etchegoyhen, David Ponce, Gorka Labaka, Nora Aranberri, Ander Corral, Xabier Saralegi, Igor Ellakuria, and Maite Martin.
\newblock 2022.
\newblock {TANDO}: A corpus for document-level machine translation.
\newblock In {\em Proceedings of the Thirteenth Language Resources and Evaluation Conference}, pages 3026--3037, Marseille, France.

\bibitem[\protect\citename{Gete \bgroup et al.\egroup }2023]{gete-etal-2023-works}
Gete, Harritxu, Thierry Etchegoyhen, and Gorka Labaka.
\newblock 2023.
\newblock What works when in context-aware neural machine translation?
\newblock In {\em Proceedings of the 24th Annual Conference of the European Association for Machine Translation}, pages 147--156, Tampere, Finland, June. European Association for Machine Translation.

\bibitem[\protect\citename{Hendy \bgroup et al.\egroup }2023]{hendy2023good}
Hendy, Amr, Mohamed Abdelrehim, Amr Sharaf, Vikas Raunak, Mohamed Gabr, Hitokazu Matsushita, Young~Jin Kim, Mohamed Afify, and Hany~Hassan Awadalla.
\newblock 2023.
\newblock How good are gpt models at machine translation? a comprehensive evaluation.

\bibitem[\protect\citename{Huang \bgroup et al.\egroup }2023]{huang2023does}
Huang, Zhihong, Longyue Wang, Siyou Liu, and Derek~F. Wong.
\newblock 2023.
\newblock How does pretraining improve discourse-aware translation?

\bibitem[\protect\citename{Huo \bgroup et al.\egroup }2020]{huo-etal-2020-diving}
Huo, Jingjing, Christian Herold, Yingbo Gao, Leonard Dahlmann, Shahram Khadivi, and Hermann Ney.
\newblock 2020.
\newblock Diving deep into context-aware neural machine translation.
\newblock In {\em Proceedings of the Fifth Conference on Machine Translation}, pages 604--616, Online, November. Association for Computational Linguistics.

\bibitem[\protect\citename{Jean \bgroup et al.\egroup }2017]{jean-etal-2017-neural}
Jean, Sebastien, Stanislas Lauly, Orhan Firat, and Kyunghyun Cho.
\newblock 2017.
\newblock Neural machine translation for cross-lingual pronoun prediction.
\newblock In {\em Proceedings of the Third Workshop on Discourse in Machine Translation}, pages 54--57, Copenhagen, Denmark, September. Association for Computational Linguistics.

\bibitem[\protect\citename{Jin \bgroup et al.\egroup }2023]{jin-etal-2023-challenges}
Jin, Linghao, Jacqueline He, Jonathan May, and Xuezhe Ma.
\newblock 2023.
\newblock Challenges in context-aware neural machine translation.
\newblock In Bouamor, Houda, Juan Pino, and Kalika Bali, editors, {\em Proceedings of the 2023 Conference on Empirical Methods in Natural Language Processing}, pages 15246--15263, Singapore, December. Association for Computational Linguistics.

\bibitem[\protect\citename{Junczys-Dowmunt \bgroup et al.\egroup }2018]{junczys-dowmunt-etal-2018-marian}
Junczys-Dowmunt, Marcin, Roman Grundkiewicz, Tomasz Dwojak, Hieu Hoang, Kenneth Heafield, Tom Neckermann, Frank Seide, Ulrich Germann, Alham~Fikri Aji, Nikolay Bogoychev, Andr{\'e} F.~T. Martins, and Alexandra Birch.
\newblock 2018.
\newblock {M}arian: Fast neural machine translation in {C}++.
\newblock In {\em Proceedings of {ACL} 2018, System Demonstrations}, pages 116--121, Melbourne, Australia, July. Association for Computational Linguistics.

\bibitem[\protect\citename{Junczys-Dowmunt}2019]{junczys-dowmunt-2019-microsoft}
Junczys-Dowmunt, Marcin.
\newblock 2019.
\newblock {M}icrosoft translator at {WMT} 2019: Towards large-scale document-level neural machine translation.
\newblock In {\em Proceedings of the Fourth Conference on Machine Translation (Volume 2: Shared Task Papers, Day 1)}, pages 225--233, Florence, Italy, August. Association for Computational Linguistics.

\bibitem[\protect\citename{Karpinska and Iyyer}2023]{karpinska2023large}
Karpinska, Marzena and Mohit Iyyer.
\newblock 2023.
\newblock Large language models effectively leverage document-level context for literary translation, but critical errors persist.

\bibitem[\protect\citename{Koehn \bgroup et al.\egroup }2007]{koehn2007moses}
Koehn, Philipp, Hieu Hoang, Alexandra Birch, Chris Callison-Burch, Marcello Federico, Nicola Bertoldi, Brooke Cowan, Wade Shen, Christine Moran, Richard Zens, et~al.
\newblock 2007.
\newblock {Moses: Open source toolkit for statistical machine translation}.
\newblock In {\em Proceedings of the 45th Annual Meeting of the Association for Computational Linguistics}, pages 177--180, Prague, Czech Republic.

\bibitem[\protect\citename{Koehn}2004]{koehn-2004-statistical}
Koehn, Philipp.
\newblock 2004.
\newblock Statistical significance tests for machine translation evaluation.
\newblock In {\em Proceedings of the 2004 Conference on Empirical Methods in Natural Language Processing}, pages 388--395, Barcelona, Spain, July. Association for Computational Linguistics.

\bibitem[\protect\citename{L{\"a}ubli \bgroup et al.\egroup }2018]{laubli-etal-2018-machine}
L{\"a}ubli, Samuel, Rico Sennrich, and Martin Volk.
\newblock 2018.
\newblock Has machine translation achieved human parity? a case for document-level evaluation.
\newblock In {\em Proceedings of the 2018 Conference on Empirical Methods in Natural Language Processing}, pages 4791--4796, Brussels, Belgium, October-November. Association for Computational Linguistics.

\bibitem[\protect\citename{Li \bgroup et al.\egroup }2020]{li-etal-2020-multi-encoder}
Li, Bei, Hui Liu, Ziyang Wang, Yufan Jiang, Tong Xiao, Jingbo Zhu, Tongran Liu, and Changliang Li.
\newblock 2020.
\newblock Does multi-encoder help? a case study on context-aware neural machine translation.
\newblock In {\em Proceedings of the 58th Annual Meeting of the Association for Computational Linguistics}, pages 3512--3518, Online, July. Association for Computational Linguistics.

\bibitem[\protect\citename{Lison \bgroup et al.\egroup }2018]{lison-etal-2018-opensubtitles2018}
Lison, Pierre, J{\"o}rg Tiedemann, and Milen Kouylekov.
\newblock 2018.
\newblock {O}pen{S}ubtitles2018: Statistical rescoring of sentence alignments in large, noisy parallel corpora.
\newblock In {\em Proceedings of the Eleventh International Conference on Language Resources and Evaluation ({LREC} 2018)}, Miyazaki, Japan, May. European Language Resources Association (ELRA).

\bibitem[\protect\citename{Lopes \bgroup et al.\egroup }2020]{lopes-etal-2020-document}
Lopes, Ant{\'o}nio, M.~Amin Farajian, Rachel Bawden, Michael Zhang, and Andr{\'e} F.~T. Martins.
\newblock 2020.
\newblock Document-level neural {MT}: A systematic comparison.
\newblock In {\em Proceedings of the 22nd Annual Conference of the European Association for Machine Translation}, pages 225--234, Lisboa, Portugal, November. European Association for Machine Translation.

\bibitem[\protect\citename{Lupo \bgroup et al.\egroup }2022]{lupo-etal-2022-focused}
Lupo, Lorenzo, Marco Dinarelli, and Laurent Besacier.
\newblock 2022.
\newblock Focused concatenation for context-aware neural machine translation.
\newblock In {\em Proceedings of the Seventh Conference on Machine Translation (WMT)}, pages 830--842, Abu Dhabi, United Arab Emirates (Hybrid), December.

\bibitem[\protect\citename{Lupo \bgroup et al.\egroup }2023]{lupo-etal-2023-encoding}
Lupo, Lorenzo, Marco Dinarelli, and Laurent Besacier.
\newblock 2023.
\newblock Encoding sentence position in context-aware neural machine translation with concatenation.
\newblock In {\em The Fourth Workshop on Insights from Negative Results in NLP}, pages 33--44, Dubrovnik, Croatia, May.

\bibitem[\protect\citename{Majumder \bgroup et al.\egroup }2022]{majumder2022baseline}
Majumder, Suvodeep, Stanislas Lauly, Maria Nadejde, Marcello Federico, and Georgiana Dinu.
\newblock 2022.
\newblock A baseline revisited: Pushing the limits of multi-segment models for context-aware translation.

\bibitem[\protect\citename{Mansimov \bgroup et al.\egroup }2021]{mansimov-etal-2021-capturing}
Mansimov, Elman, G{\'a}bor Melis, and Lei Yu.
\newblock 2021.
\newblock Capturing document context inside sentence-level neural machine translation models with self-training.
\newblock In {\em Proceedings of the 2nd Workshop on Computational Approaches to Discourse}, pages 143--153, Punta Cana, Dominican Republic and Online, November. Association for Computational Linguistics.

\bibitem[\protect\citename{Mcnemar}1947]{Mcnemar1947NoteOT}
Mcnemar, Quinn.
\newblock 1947.
\newblock Note on the sampling error of the difference between correlated proportions or percentages.
\newblock {\em Psychometrika}, 12:153--157.

\bibitem[\protect\citename{M{\"u}ller \bgroup et al.\egroup }2018]{muller-etal-2018-large}
M{\"u}ller, Mathias, Annette Rios, Elena Voita, and Rico Sennrich.
\newblock 2018.
\newblock A large-scale test set for the evaluation of context-aware pronoun translation in neural machine translation.
\newblock In {\em Proceedings of the Third Conference on Machine Translation: Research Papers}, pages 61--72, Brussels, Belgium, October. Association for Computational Linguistics.

\bibitem[\protect\citename{Nagata and Morishita}2020]{nagata-morishita-2020-test}
Nagata, Masaaki and Makoto Morishita.
\newblock 2020.
\newblock A test set for discourse translation from {J}apanese to {E}nglish.
\newblock In {\em Proceedings of the 12th Language Resources and Evaluation Conference}, pages 3704--3709, Marseille, France, May. European Language Resources Association.

\bibitem[\protect\citename{Papineni \bgroup et al.\egroup }2002]{papineni-etal-2002-bleu}
Papineni, Kishore, Salim Roukos, Todd Ward, and Wei-Jing Zhu.
\newblock 2002.
\newblock {B}leu: a method for automatic evaluation of machine translation.
\newblock In {\em Proceedings of the 40th Annual Meeting of the Association for Computational Linguistics}, pages 311--318, Philadelphia, Pennsylvania, USA, July. Association for Computational Linguistics.

\bibitem[\protect\citename{Petrick \bgroup et al.\egroup }2023]{petrick-etal-2023-document}
Petrick, Frithjof, Christian Herold, Pavel Petrushkov, Shahram Khadivi, and Hermann Ney.
\newblock 2023.
\newblock Document-level language models for machine translation.
\newblock In Koehn, Philipp, Barry Haddow, Tom Kocmi, and Christof Monz, editors, {\em Proceedings of the Eighth Conference on Machine Translation}, pages 375--391, Singapore, December. Association for Computational Linguistics.

\bibitem[\protect\citename{Post and Junczys-Dowmunt}2023]{post2023escaping}
Post, Matt and Marcin Junczys-Dowmunt.
\newblock 2023.
\newblock Escaping the sentence-level paradigm in machine translation.
\newblock {\em arXiv preprint arXiv:2304.12959v1}.

\bibitem[\protect\citename{Post}2018]{post-2018-call}
Post, Matt.
\newblock 2018.
\newblock A call for clarity in reporting {BLEU} scores.
\newblock In {\em Proceedings of the Third Conference on Machine Translation: Research Papers}, pages 186--191, Brussels, Belgium, October. Association for Computational Linguistics.

\bibitem[\protect\citename{Scherrer \bgroup et al.\egroup }2019]{scherrer-etal-2019-analysing}
Scherrer, Yves, J{\"o}rg Tiedemann, and Sharid Lo{\'a}iciga.
\newblock 2019.
\newblock Analysing concatenation approaches to document-level {NMT} in two different domains.
\newblock In {\em Proceedings of the Fourth Workshop on Discourse in Machine Translation (DiscoMT 2019)}, pages 51--61, Hong Kong, China, November. Association for Computational Linguistics.

\bibitem[\protect\citename{Sennrich \bgroup et al.\egroup }2016]{sennrich2016bpe}
Sennrich, Rico, Barry Haddow, and Alexandra Birch.
\newblock 2016.
\newblock Neural machine translation of rare words with subword units.
\newblock In {\em Proceedings of the 54th Annual Meeting of the Association for Computational Linguistics (Volume 1: Long Papers)}, pages 1715--1725, Berlin, Germany, August. Association for Computational Linguistics.

\bibitem[\protect\citename{Sugiyama and Yoshinaga}2019]{sugiyama-yoshinaga-2019-data}
Sugiyama, Amane and Naoki Yoshinaga.
\newblock 2019.
\newblock Data augmentation using back-translation for context-aware neural machine translation.
\newblock In {\em Proceedings of the Fourth Workshop on Discourse in Machine Translation (DiscoMT 2019)}, pages 35--44, Hong Kong, China, November. Association for Computational Linguistics.

\bibitem[\protect\citename{Sun \bgroup et al.\egroup }2022]{sun-etal-2022-rethinking}
Sun, Zewei, Mingxuan Wang, Hao Zhou, Chengqi Zhao, Shujian Huang, Jiajun Chen, and Lei Li.
\newblock 2022.
\newblock Rethinking document-level neural machine translation.
\newblock In {\em Findings of the Association for Computational Linguistics: ACL 2022}, pages 3537--3548, Dublin, Ireland, May. Association for Computational Linguistics.

\bibitem[\protect\citename{Sutskever \bgroup et al.\egroup }2014]{sutskever2014sequence}
Sutskever, Ilya, Oriol Vinyals, and Quoc~V Le.
\newblock 2014.
\newblock Sequence to sequence learning with neural networks.
\newblock In {\em Advances in neural information processing systems}, pages 3104--3112.

\bibitem[\protect\citename{Tiedemann and Scherrer}2017]{tiedemann-scherrer-2017-neural}
Tiedemann, J{\"o}rg and Yves Scherrer.
\newblock 2017.
\newblock Neural machine translation with extended context.
\newblock In {\em Proceedings of the Third Workshop on Discourse in Machine Translation}, pages 82--92, Copenhagen, Denmark, September. Association for Computational Linguistics.

\bibitem[\protect\citename{Vaswani \bgroup et al.\egroup }2017]{VaswaniSPUJGKP17}
Vaswani, Ashish, Noam Shazeer, Niki Parmar, Jakob Uszkoreit, Llion Jones, Aidan~N. Gomez, Lukasz Kaiser, and Illia Polosukhin.
\newblock 2017.
\newblock Attention is all you need.
\newblock In {\em Advances in Neural Information Processing Systems 30}, pages 5998--6008.

\bibitem[\protect\citename{Voita \bgroup et al.\egroup }2018]{voita-etal-2018-context}
Voita, Elena, Pavel Serdyukov, Rico Sennrich, and Ivan Titov.
\newblock 2018.
\newblock Context-aware neural machine translation learns anaphora resolution.
\newblock In {\em Proceedings of the 56th Annual Meeting of the Association for Computational Linguistics (Volume 1: Long Papers)}, pages 1264--1274, Melbourne, Australia, July. Association for Computational Linguistics.

\bibitem[\protect\citename{Voita \bgroup et al.\egroup }2019a]{voita-etal-2019-context}
Voita, Elena, Rico Sennrich, and Ivan Titov.
\newblock 2019a.
\newblock Context-aware monolingual repair for neural machine translation.
\newblock In {\em Proceedings of the 2019 Conference on Empirical Methods in Natural Language Processing and the 9th International Joint Conference on Natural Language Processing (EMNLP-IJCNLP)}, pages 877--886, Hong Kong, China, November. Association for Computational Linguistics.

\bibitem[\protect\citename{Voita \bgroup et al.\egroup }2019b]{voita-etal-2019-good}
Voita, Elena, Rico Sennrich, and Ivan Titov.
\newblock 2019b.
\newblock When a good translation is wrong in context: Context-aware machine translation improves on deixis, ellipsis, and lexical cohesion.
\newblock In {\em Proceedings of the 57th Annual Meeting of the Association for Computational Linguistics}, pages 1198--1212, Florence, Italy, July. Association for Computational Linguistics.

\bibitem[\protect\citename{Wang \bgroup et al.\egroup }2023]{wang2023documentlevel}
Wang, Longyue, Chenyang Lyu, Tianbo Ji, Zhirui Zhang, Dian Yu, Shuming Shi, and Zhaopeng Tu.
\newblock 2023.
\newblock Document-level machine translation with large language models.

\bibitem[\protect\citename{Wong and Kit}2012]{wong-kit-2012-extending}
Wong, Billy T.~M. and Chunyu Kit.
\newblock 2012.
\newblock Extending machine translation evaluation metrics with lexical cohesion to document level.
\newblock In {\em Proceedings of the 2012 Joint Conference on Empirical Methods in Natural Language Processing and Computational Natural Language Learning}, pages 1060--1068, Jeju Island, Korea, July. Association for Computational Linguistics.

\bibitem[\protect\citename{Wu \bgroup et al.\egroup }2022]{DBLP:journals/ml/WuXZWXQ22}
Wu, Xueqing, Yingce Xia, Jinhua Zhu, Lijun Wu, Shufang Xie, and Tao Qin.
\newblock 2022.
\newblock A study of {BERT} for context-aware neural machine translation.
\newblock {\em Mach. Learn.}, 111(3):917--935.

\bibitem[\protect\citename{Xu \bgroup et al.\egroup }2021]{XuLW0C21}
Xu, Mingzhou, Liangyou Li, Derek~F. Wong, Qun Liu, and Lidia~S. Chao.
\newblock 2021.
\newblock Document graph for neural machine translation.
\newblock In Moens, Marie{-}Francine, Xuanjing Huang, Lucia Specia, and Scott~Wen{-}tau Yih, editors, {\em Proceedings of the 2021 Conference on Empirical Methods in Natural Language Processing, {EMNLP} 2021, Virtual Event / Punta Cana, Dominican Republic, 7-11 November, 2021}, pages 8435--8448. Association for Computational Linguistics.

\bibitem[\protect\citename{Zhang \bgroup et al.\egroup }2018]{zhang-etal-2018-improving}
Zhang, Jiacheng, Huanbo Luan, Maosong Sun, Feifei Zhai, Jingfang Xu, Min Zhang, and Yang Liu.
\newblock 2018.
\newblock Improving the transformer translation model with document-level context.
\newblock In {\em Proceedings of the 2018 Conference on Empirical Methods in Natural Language Processing}, pages 533--542, Brussels, Belgium, October-November. Association for Computational Linguistics.

\end{thebibliography}
\bibliographystyle{eamt24}

\appendix

\end{document}